# Deep Learning for Molecular Graphs with Tiered Graph Autoencoders and Graph Prediction

Daniel T. Chang (张遵)

*IBM (Retired)* dtchang43@gmail.com

**Abstract:**

Tiered graph autoencoders provide the architecture and mechanisms for learning tiered latent representations and latent spaces for molecular graphs that explicitly represent and utilize groups (e.g., functional groups). This enables the utilization and exploration of tiered molecular latent spaces, either individually—the node (atom) tier, the group tier, or the graph (molecule) tier—or jointly, as well as navigation across the tiers. In this paper, we discuss the use of tiered graph autoencoders together with graph prediction for molecular graphs. We show features of molecular graphs used, and groups in molecular graphs identified for some sample molecules. We briefly review graph prediction and the QM9 dataset for background information, and discuss the use of tiered graph embeddings for graph prediction, particularly weighted group pooling. We find that functional groups and ring groups effectively capture and represent the chemical essence of molecular graphs (structures). Further, tiered graph autoencoders and graph prediction together provide effective, efficient and interpretable deep learning for molecular graphs, with the former providing unsupervised, transferable learning and the latter providing supervised, task-optimized learning.

## 1 Introduction

*Tiered graph autoencoders* [1-2] provide the architecture and mechanisms for learning *tiered latent representations and latent spaces* [2] for molecular graphs that explicitly represent and utilize groups (e.g., functional groups). At each tier, the latent representations consist of: node features, edge indices, edge features, membership matrix, and node embeddings. (At higher tiers, nodes and edges are coarsened.) This enables the utilization and exploration of tiered molecular latent spaces, either individually—the node (atom) tier, the group tier, or the graph (molecule) tier—or jointly, as well as navigation across the tiers. *PyTorch Geometric (PyG)* [5] is an easy-to-use but powerful library for implementing tiered graph autoencoders.

In this paper, we discuss the use of tiered graph autoencoders together with graph prediction for molecular graphs. *Graph prediction* with molecular properties or molecular activities as targets is of great academic and practical significance [3], with the former applicable to QSPRs [3] and the latter applicable to QSARs [3, 6]. It is also critical to molecular design [3, 7] since the new molecules must possess desired molecular properties / activities.

Tiered graph autoencoders have been discussed in detail in [1-2]. Therefore, we only show features of molecular graphs used, and groups in molecular graphs identified for some sample molecules, in what follows. The following databases / datasets are accessed for sample molecules: *PubChem* [1, 8-10] / *ALATIS* [1, 11], *QM9* [12-13] and *BindingDB* [14]. All of these are accessible as *SDF (Structure Data File)* [1, 15] files. The QM9 dataset (accessed through PyG) has 133K molecules and the BindingDB dataset (drawn from PubChem) has 121K molecules.

We briefly review graph prediction and the QM9 dataset for background information, and discuss the use of *tiered graph embeddings* for graph prediction, particularly *weighted group pooling*. To learn and optimize *group weights* for particular molecular properties / activities, we briefly discuss two approaches: the *weighted-sum pooling* approach where group weights function as model parameters and the *hyperparameter optimization* approach where group weights are treated as hyperparameters.

## 2 Unsupervised Deep Learning: Tiered Graph Autoencoders

The following diagram shows the *tiered graph autoencoder architecture* [2] for learning tiered latent representations and latent spaces:

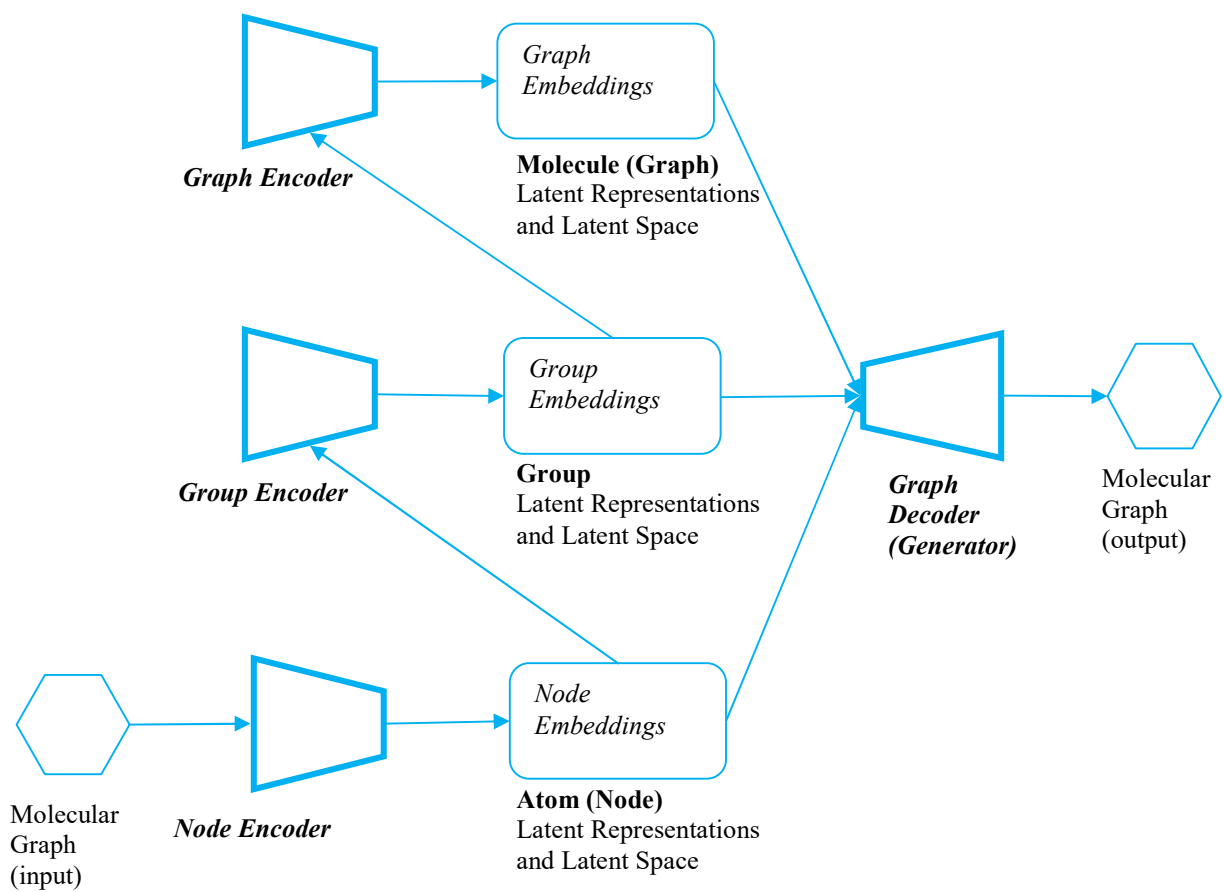

*Tiered graph autoencoders* have been discussed in detail in [1-2]. In the following, therefore, we only show features of molecular graphs used, and groups in molecular graphs identified for some sample molecules. The following databases / datasets are accessed for sample molecules: PubChem / ALATIS, QM9 and BindingDB. All of these are accessible as SDF files. *Note:* the QM9 and BindingDB datasets do not support ALATIS unique atom identifiers [1, 11].

## 2.1 Features of Molecular Graphs

### Molecule (Graph) Features

The following table shows the molecule (graph) features used and their information sources. The PubChem CID, when available, is very useful for identifying a molecular graph and its tiered embeddings as well as for relating these with the actual molecule and its structure.

| *Feature* | *PubChem* | *RDKit [16]* | *Other [1, 17]* |
|---|---|---|---|
| CID | 'PUBCHEM_COMPOUND_CID' or 'PubChem CID' | Mol.GetProp() | |
| Functional groups | | | identify_functional_groups() |
| Rings | | Mol.GetRingInfo() | |

### Atom (Node) Features

The following table shows the atom (node) features used and their information sources. As can be seen, the two feature sets used are different but are identical in the most important features: atomic type ('H', 'C', 'N', 'O', or 'F'), hybridization (SP, SP2, or SP3) and aromaticity. To generate and use transferable atom (node) embeddings, however, there need to be a widely agreed-upon atom (node) feature set.

| *Feature* | *RDKit* | *mol_to_graph()* (based on mol2vec() [1], applied to BindingDB) | *QM9 (PyG)* |
|---|---|---|---|
| Atom type | Atom.GetSymbol() | 44 (one-hot) (**11 reactive nonmetal**)<br>['**C**', '**N**', '**O**', '**S**', '**F**', '**Si**', '**P**', '**Cl**', '**Br**', 'Mg', 'Na', 'Ca', 'Fe', 'As', 'Al', '**I**', 'B', 'V', 'K', 'Tl', | 5 (one-hot)<br>['H', 'C', 'N', 'O', 'F'] |

| | | 'Yb', 'Sb', 'Sn', 'Ag', 'Pd', 'Co', 'Se', 'Ti', 'Zn', '**H**', 'Li', 'Ge', 'Cu', 'Au', 'Ni', 'Cd', 'In', 'Mn', 'Zr', 'Cr', 'Pt', 'Hg', 'Pb', 'unknown'] | |
|---|---|---|---|
| Degree | Atom.GetDegree() | 11 (one-hot) [0, 1, 2, 3, 4, 5, 6, 7, 8, 9, 10] | - |
| Implicit valence | Atom.GetImplicitValence() | 7 (one-hot) [0, 1, 2, 3, 4, 5, 6] | - |
| Formal charge | Atom.GetFormalCharge() | 1 | - |
| Radical Electrons | Atom.GetNumRadicalElectrons() | 1 | - |
| Hybridization | Atom.GetHybridization() | 5 (one-hot) [SP, SP2, SP3, SP3D, SP3D2] | 3 (one-hot) [SP, SP2, SP3] |
| Aromaticity | Atom.GetIsAromatic() | 1 | 1 |
| Atomic number | Atom.GetAtomicNum() | - | 1 |
| Number of hydrogens | Atom.GetTotalNumHs() | - | 1 |
| Donor | ChemicalFeatures | - | 1 |
| Acceptor | ChemicalFeatures | - | 1 |

## Bond (Edge) Features

The following table shows the bond (edge) features used and their information sources. As can be seen, the two feature sets used are different but are identical in the most important feature: bond type (SINGLE, DOUBLE, TRIPLE, or AROMATIC).

| Feature | RDKit | mol_to_graph() (based on mol2vec(), applied to BindingDB) | QM9 (PyG) |
|---|---|---|---|
| Bond type | Bond.GetBondType() | 4 (one-hot) [SINGLE, DOUBLE, TRIPLE, AROMATIC] | 4 (one-hot) [SINGLE, DOUBLE, TRIPLE, AROMATIC] |
| Conjugated bond | Bond. GetIsConjugated() | 1 | - |
| Ring | Bond.IsInRing() | 1 | - |

## 2.2 Groups in Molecular Graphs

The following table shows groups in molecular graphs identified for some sample molecules, listed in the order of the number of constituent atoms, with atom index (starting at 0) based on the ALATIS unique atom identifier (starting at 1). The types of groups [2] include: *FG (functional group)*, *RG (ring group)* and *CCG (connected-component group)*. Note that the RG is more general than the AG (aromatic ring group) discussed in [2] and it does not include attached hydrogen atoms.

As can be seen from the table, connected components (not belonging to either FGs or RGs) contain hydrogen atoms and, occasionally, carbon atoms. We group them into a single CCG since they do not have much chemical significance. *Functional groups and ring groups* effectively capture and represent the chemical essence of molecular graphs (structures).

| Compound (cid) | Atoms (index) | FGs | RGs | CCG |
|---|---|---|---|---|
| formaldehyde (712) | C(0) O(1) H(2:3) | (0, 1) | - | (2, 3) |
| cyanogen (9999) | C(0:1) N (2:3) | (0, 1, 2, 3) | - | - |
| methane (297) | C(0) H (1:4) | - | - | (0, 1, 2, 3, 4) |
| carbonic acid (767) | C(0) O(1:3) H(4:5) | (0, 1, 2, 3) | - | (4, 5) |
| acetaldehyde (177) | C(0: 1) O(2) H(3:6) | (1, 2) | - | (0, 3, 4, 5, 6) |
| ethanol (702) | C(0:1) O(2) H(3:8) | (2,) | - | (0, 1, 3, 4, 5, 6, 7, 8) |
| bezene (11309472) | C(0:5) H(6:11) | - | (0, 2, 4, 5, 3, 1) | (6, 7, 8, 9, 10, 11) |
| methoxyethane (10903) | C(0:2) O(3) H(4:11) | (3,) | - | (0, 1, 2, 4, 5, 6, 7, 8, 9, 10, 11) |
| methoxybenzene (7519) | C(0:6) O(7) H(8:15) | (7,) | (1, 3, 5, 6, 4, 2) | (0, 8, 9, 10, 11, 12, 13, 14, 15) |
| vanillin (1183) | C(0:7) O(8:10) H(11:18) | ((8, 4), (9,), (10,)) | (1, 5, 3, 7, 6, 2) | (0, 11, 12, 13, 14, 15, 16, 17, 18) |
| mesitylene (7947) | C(0:8) H(9:20) | - | (3, 7, 5, 8, 4, 6) | (0, 1, 2, 9, 10, 11, 12, 13, 14, 15, 16, 17, 18, 19, 20) |
| 657862 | C(0:11) N(12:13) O(14:15) S(16:17) H(18:31) | ((12,), (13,), (14,), (15,), (16,), (17,)) | ((3, 11, 16, 4, 6, 5), (8, 12, 10, 13, 9, 7), (17, 6, 5, 7, 9)) | (0, 1, 2, 18, 19, 20, 21, 22, 23, 24, 25, 26, 27, 28, 29, 30, 31) |
| 652912 | C(0:13) N(14) O(15:17) H(18:32) | ((1, 3), (16, 13, 15), (14,), (17,)) | ((1, 3, 7, 9, 4), (2, 5, 8, 11, 10, 6), (12, 14, 11, 8, 7, 9)) | (0, 18, 19, 20, 21, 22, 23, 24, 25, 26, 27, 28, 29, 30, 31, 32) |
| 649963 | C(0:14) N(15:17) O(18) H(19:33) | ((9, 15), (16,), (17,), (18,)) | ((1, 2, 6, 17, 5), (3, 12, 8, 10, 13, 4), (7, 11, 14, 16, 13, 10)) | (0, 19, 20, 21, 22, 23, 24, 25, 26, 27, 28, 29, 30, 31, 32, 33) |
| 656318 | C(0:21) N(22:25) O(26:29) S(30) H(31:46) | ((22,), (23,), (24,), (25,), (26,), (27, 28, 30), (29,)) | ((0, 2, 8, 15, 7, 1), (3, 9, 18, 24, 10, 4), (5, 11, 29, 14, 6), (12, 17, 19, 25, 20, 16), (21, 24, 18, 23, 20, 16)) | (13, 31, 32, 33, 34, 35, 36, 37, 38, 39, 40, 41, 42, 43, 44, 45, 46) |
| 644735 | C(0:24) N(25:29) O(30:32) H(33:53) | ((32,), (26, 30, 23), (25,), (27,), (28,), (29,), (31,)) | ((1, 5, 20, 28, 10, 2), (3, 11, 32, 17, 4), (6, 15, 7, 9, 16, 8), (12, 19, 22, 29, 21, 18), (24, 28, 20, 27, 22, 19)) | (0, 13, 14, 33, 34, 35, 36, 37, 38, 39, 40, 41, 42, 43, 44, 45, 46, 47, 48, 49, 50, 51, 52, 53) |
| 657445 | C(0:24) N(25:30) H(31:58) | ((26, 14), (25,), (27,), (28,), (29,), (30,)) | ((0, 3, 7, 18, 6, 2), (1, 5, 9, 19, 8, 4), (10, 13, 30, 17, 22, 20), (21, 23, 29, 24, 22, 20)) | (11, 12, 15, 16, 31, 32, 33, 34, 35, 36, 37, 38, 39, 40, 41, 42, 43, 44, 45, 46, 47, 48, 49, 50, 51, 52, 53, 54, 55, 56, 57, 58) |

## 3 Supervised Deep Learning: Graph Prediction

*Graph prediction* with molecular properties or molecular activities as targets is of great academic and practical significance, with the formal applicable to QSPRs and the latter applicable to QSARs. It is also critical to molecular design since the new molecules must possess desired molecular properties / activities. In the following, we briefly review graph prediction and the QM9 dataset for background information, and discuss the use of *tiered graph embeddings* for graph prediction, particularly *weighted group pooling*.

### 3.1 Synopsis of Graph Prediction

For *graph prediction* tasks [18], to obtain a compact embedding on the graph level, a *pooling layer* is used to coarsen a graph into sub-graphs or to sum/average over the node embeddings. In the case of using *GCNs (graph convolutional networks)* [19] for graph prediction, a GCN layer is followed by a pooling layer to coarsen a graph into sub-graphs so that node embeddings on coarsened graphs represent higher-level embeddings. Multiple GCN-pooling layers may be used. To calculate the probability for each graph label / target, the output layer is a *MLP (multilayer perceptron)* with the SoftMax function.

A major limitation of most GCN-pooling architectures is that they are inherently flat as they only propagate information across the edges of the graph and are unable to infer and aggregate the information in a hierarchical way [20]. This *lack of hierarchical structure* is especially problematic for graph prediction. In order to successfully encode molecular graphs for the task of graph prediction, one would ideally want to encode the *local molecular structure* (e.g., individual atoms and their direct bonds) as well as the *coarse-grained molecular structure* (e.g., groups of atoms and bonds representing functional groups) of the molecular graph.

*DiffPool* [20] is a differentiable graph pooling module that can generate *hierarchical representations of graphs*. It learns a differentiable soft cluster assignment for nodes at each layer of a deep GNN (Graph Neural Network), mapping nodes to a set of *clusters*, which then form the coarsened input for the next GNN layer. At each hierarchical layer, it runs a GNN model to obtain *embeddings* of nodes. It then uses these learned embeddings to cluster nodes together and run another GNN layer on this coarsened graph. DiffPool achieves good results. However, it is not clear at all how well the clusters learned match the functional groups that they purport to represent. Also, DiffPool has high computational cost due to the (double) learning of clustering and embeddings at each layer.

### 3.2 QM9

*QM9* [13, 21] is a widely used prediction dataset that provides geometric, energetic, electronic and thermodynamic properties for a subset of the GDB-17 database, comprising 133K drug-like organic molecules (as accessed through PyG). It has *12 target molecular properties*:

- mu - dipole moment (unit: Debye)
- alpha - isotropic polarizability (unit: Bohr$^3$)

- HOMO - highest occupied molecular orbital energy (unit: Hartree)
- LUMO - lowest unoccupied molecular orbital energy (unit: Hartree)
- gap - gap between HOMO and LUMO (unit: Hartree)
- $R^2$ - electronic spatial extent (unit: $Bohr^2$)
- ZPVE - zero point vibrational energy (unit: Hartree)
- $U_0$ - internal energy at 0K (unit: Hartree)
- U - internal energy at 298.15K (unit: Hartree)
- H - enthalpy at 298.15K (unit: Hartree)
- G - free energy at 298.15K (unit: Hartree)
- Cv - heat capacity at 298.15K (unit: cal/(mol*K))

Among these, three properties (mu, alpha and $R^2$) measure the spatial distribution of electrons in the molecule, three properties (HOMO. LUMO and gap) concern the state of electrons in the molecule, four properties ($U_0$, U, H, and G) relate to the atomization energy, and the rest (ZPVE and Cv) relate to the fundamental vibration and the thermodynamic property, respectively.

QM9 is supported in PyG as an in-memory dataset. To use it for our work, we extend it to support the *identification of groups* (FG, RG and CCG) as well as the *generation of group and graph membership matrices* for molecular graphs.

## 3.3 Graph Prediction Using Tiered Graph Embeddings

*Tiered graph autoencoders* [1-2] provide the most direct and effective architecture and mechanisms for generating *hierarchical representations (embeddings) of molecular graphs* since they are based on the direct representation and utilization of *groups*, the chemically most significant hierarchical components of molecular graphs. It is therefore natural and advantageous to use *tiered graph embeddings*, generated by tiered graph autoencoders, for molecular graph prediction, as shown below:

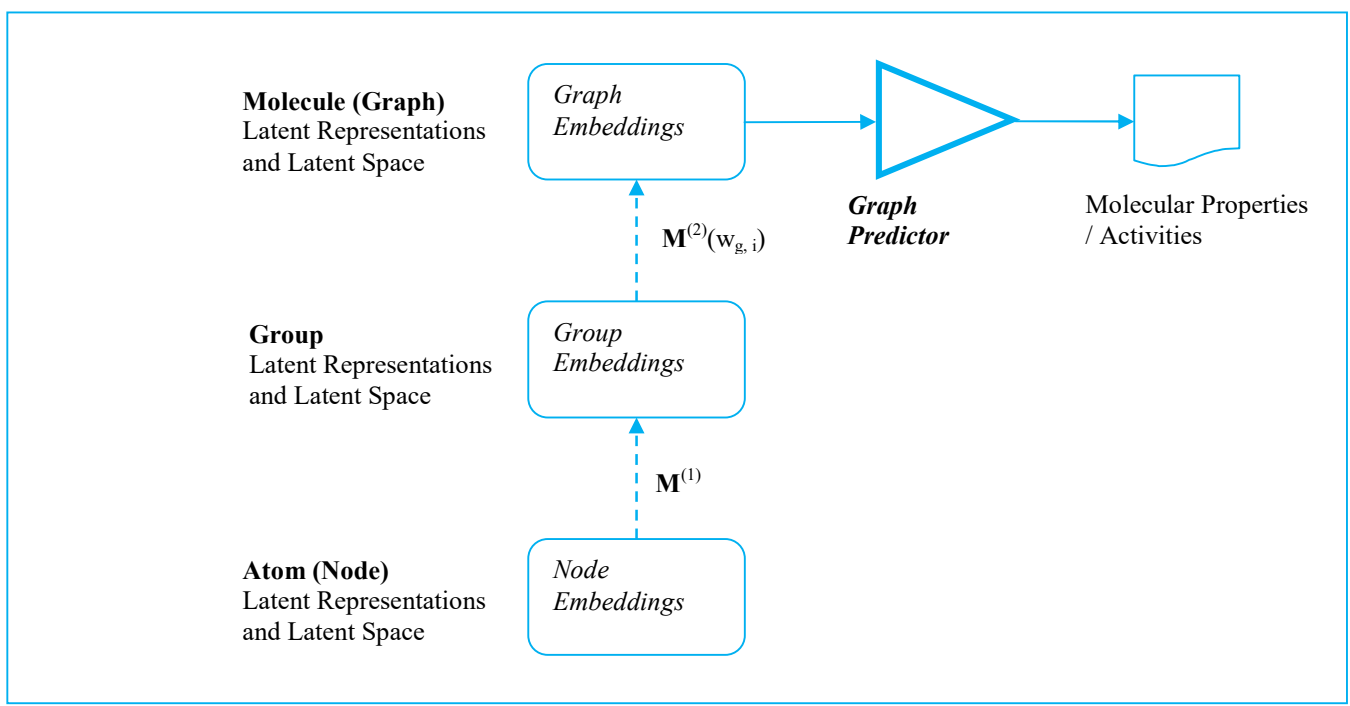

In the diagram, $\mathbf{M}^{(1)}$ is the *group membership matrix*. Based on *DiffGroupPool* [1-2], we have

$$\mathbf{X}^{(2)} = (\mathbf{M}^{(1)})^T \mathbf{Z}^{(1)}$$

where $\mathbf{Z}^{(1)}$ is the node embeddings and $\mathbf{X}^{(2)}$ is the initial group embeddings. Further, $\mathbf{M}^{(2)}$ is the *graph membership matrix*. Based on DiffGroupPool, we have

$$\mathbf{X}^{(3)} = (\mathbf{M}^{(2)})^T \mathbf{Z}^{(2)}$$

where $\mathbf{Z}^{(2)}$ is the group embeddings and $\mathbf{X}^{(3)}$ is the initial graph embeddings. $w_{g,i}$ is the *group weight* for group i in the graph membership matrix $\mathbf{M}^{(2)}$, to be discussed below.

Since graph embeddings are in the proper form (convoluted and pooled from node embeddings and subsequently group embeddings) for use as the input layer for graph prediction, the *graph predictor* is a simple *MLP* with two hidden layers.

### 3.4 Weighted Group Pooling

Whereas in pure tiered graph autoencoders [1-2] all groups in the graph membership matrix $\mathbf{M}^{(2)}$ have an equal weight of 1, to use tiered graph autoencoders with graph prediction we introduce the *group weight* $w_{g,i}$ to represent and account for the different importance of group i for predicting different target molecular property / activity. For proof of concept we use the following *constant* group weights:

- $w_{g,i} = 1$ if i is a FG,
- $w_{g,i} = 0.5$ if i is a RG,
- $w_{g,i} = 0.1$ if i is a CCG.

This assumes FGs are most important, RGs are less important, and a CCG is the least important for predicting molecular properties / activities.

For comprehensive studies, however, we foresee the use of *fine-grain (variable) group weights*, with different group weights for different types of FGs and RGs (see 2.2 Groups in Molecular Graphs) and with group weights to be different for different target molecular properties / activities. This will allow us to *learn and optimize* group weights based on the results of graph prediction learning.

Using (variable) group weights $w_{g,i}$ in the graph membership matrix $\mathbf{M}^{(2)}$ amounts to *weighted group pooling*. Weighted group pooling thus serves as the *link / coupling* between (unsupervised) tiered graph autoencoders and (supervised) graph prediction, enabling *task-independent node and group embeddings*, generated using tiered graph autoencoders, to be used to build *task-optimized graph embeddings*, generated using graph prediction for specific targets.

### Weighted-Sum Pooling

For weighted group pooling, it should be possible to *learn* group weights as part of graph prediction, with joint learning of group pooling and graph prediction, similar to weighted geometric pooling [22] used for image prediction.

In this approach, we will use *group embeddings*, generated using tiered graph autoencoders up to Tier 2, as input for graph prediction. We will need a *weighted-sum pooling layer*, which is not supported in PyG or PyTorch as far as we know, to use before the MLP layer. The group weights thus function as *model parameters*. To improve performance, the group embeddings can be *saved and loaded* for use in graph prediction tasks.

We will explore this further in the future.

**Hyperparameter Optimization**

Alternatively, for weighted group pooling, we can treat group weights as *hyperparameters* and use *hyperparameter optimization* [23-24] to learn and optimize group weights. Learning of group pooling is separate from learning of graph prediction, but depends on the latter's results.

In this approach, we will use *graph embeddings*, generated using tiered graph autoencoders up to Tier 3, as input for graph prediction. We will need to (1) select a *strategy* (e.g., random search, Bayesian Optimization) for hyperparameters (group weights) search, (2) based on the strategy, design a *set, or sets, of group weights* to use, (3) generate *graph embeddings* for the designed set(s) of group weights, (4) perform a *target-specific graph prediction learning* for each generated graph embedding, (5) determine *the optimal set of group weights* that produce the best graph prediction results for the target, and (6) repeat from step 2, if the strategy calls for, until a *termination condition* is reached. To improve performance, the graph embeddings, generated for the designed sets of group weights, can be *saved and loaded* for use for other graph prediction targets, bypassing step 3.

We will explore this further in the future.

## 4 Summary and Conclusion

In this paper we discussed the use of tiered graph autoencoders together with graph prediction for molecular graphs. Graph prediction with molecular properties / activities as targets is of great academic and practical significance. It is also critical to molecular design. PyG with PyTorch prove to be an easy-to-use but powerful library for implementing tiered graph autoencoders together with graph prediction.

We showed features of molecular graphs used, and groups in molecular graphs identified for some sample molecules. We briefly reviewed graph prediction and the QM9 dataset for background information, and discussed the use of tiered graph embeddings for graph prediction, particularly weighted group pooling.

We find that functional groups and ring groups effectively capture and represent the chemical essence of molecular graphs (structures). For example, Gallotannic acid has 174 atoms and a very complex molecular structure. However, its essence is effectively represented by 35 FGs and 11 RGs, with 1 CCG containing the remaining 52 hydrogen atoms and 1 carbon atom. In all, for the QM9 dataset (accessed through PyG, 133K molecules) a molecule has on average 4.79 groups, with a minimum of 1 group and a maximum of 11 groups; for the BindingDB dataset (drawn from PubChem, 121K molecules) a molecule has on average 8.97 groups, with a minimum of 2 groups and a maximum of 47 groups (Gallotannic Acid).

Further, we find that tiered graph autoencoders and graph prediction together provide effective, efficient and interpretable deep learning for molecular graphs, with the former providing unsupervised, transferable learning and the latter providing supervised, task-optimized learning. It is effective for generating and utilizing hierarchical representations (embeddings) of molecular graphs since tiered graph autoencoders are based on the direct representation and utilization of groups, the chemically most significant hierarchical components of molecular graphs. It is efficient since tiered graph autoencoders learning and graph prediction learning are essentially separated, coupled only through weighted group pooling. Group or graph embeddings, generated by tiered graph autoencoders, can be saved and reloaded for use in graph prediction tasks. It is interpretable since tiered graph autoencoders are based on chemical concepts: atom (node), bond (edge), group and molecule (graph) as well as atom features and bond features. The graph prediction targets are chemical concepts: molecular properties / activities, and even group weights are based on the concept of chemical functionality / reactivity of groups. Being concept-oriented deep learning [4], it is feasible to understand and interpret deep learning results.

For future work, we plan to explore weighed group pooling for molecular graph prediction, investigating both the weighted-sum pooling approach, where group weights function as model parameters, and the hyperparameter optimization approach, where group weights are treated as hyperparameters. We also plan to utilize / extend tiered graph autoencoders with graph prediction for molecular design, which involves molecular graph generation using task-optimized tiered embeddings (graph, group and node) and membership matrices.